\documentclass[10pt,twocolumn,letterpaper]{article}

\usepackage{iccv}
\usepackage{times}
\usepackage{epsfig}
\usepackage{graphicx}
\usepackage{amsmath}
\usepackage{amssymb}
\usepackage{helvet}
\usepackage{courier}
\usepackage{color}
\usepackage{comment}
\usepackage{amsmath,amssymb,amsfonts}
\usepackage{algorithmic}
\usepackage{textcomp}
\usepackage{xcolor}
\usepackage{epsfig}
\usepackage{algorithm}
\usepackage{algorithmic}
\usepackage{soul}
\usepackage{cool}
\usepackage{nameref}
\usepackage{float}
\usepackage[utf8x]{inputenc} 
\usepackage[breaklinks=true,bookmarks=false]{hyperref}
\usepackage{enumitem}

\ificcvfinal\fi

\begin{document}

\iccvfinalcopy
\title{Fusion-FlowNet: Energy-Efficient Optical Flow Estimation using Sensor Fusion and Deep Fused Spiking-Analog Network Architectures}

\author{Chankyu Lee, Adarsh Kumar Kosta and Kaushik Roy\\
Purdue University, West Lafayette, IN 47907, USA\\
{\tt\small \{lee2216, akosta, kaushik\}@purdue.edu}
}

\maketitle
\ificcvfinal\thispagestyle{empty}\fi

\begin{abstract} \label{abstract}

Standard frame-based cameras that sample light intensity frames are heavily impacted by motion blur for high-speed motion and fail to perceive scene accurately when the dynamic range is high. Event-based cameras, on the other hand, overcome these limitations by asynchronously detecting the variation in individual pixel intensities. However, event cameras only provide information about pixels in motion, leading to sparse data. Hence, estimating the overall dense behavior of pixels is difficult. To address such issues associated with the sensors, we present Fusion-FlowNet, a sensor fusion framework for energy-efficient optical flow estimation using both frame- and event-based sensors, leveraging their complementary characteristics. Our proposed network architecture is also a fusion of Spiking Neural Networks (SNNs) and Analog Neural Networks (ANNs) where each network is designed to simultaneously process asynchronous event streams and regular frame-based images, respectively. Our network is end-to-end trained using unsupervised learning to avoid expensive video annotations. The method generalizes well across distinct environments (rapid motion and challenging lighting conditions) and demonstrates state-of-the-art optical flow prediction on the Multi-Vehicle Stereo Event Camera (MVSEC) dataset. Furthermore, our network offers substantial savings in terms of the number of network parameters and computational energy cost.

\end{abstract}

\section{Introduction} \label{sec:intro}
Optical flow estimation is a fundamental computer vision problem, allowing us to visualize the motion field in scenes. It involves estimating the spatio-temporal motion patterns of pixels and forms the groundwork for more complex tasks such as motion segmentation \cite{narayana2013coherent} and action recognition \cite{wang2011action}. Over the past years, the optical flow estimation has been largely dominated by conventional computer vision algorithms such as differential \cite{lucaskanade}, phase correlation and block-based methods \cite{beauchemin1995computation}. Recently, deep Analog Neural Networks
(ANNs\footnote[1]{We refer to the conventional deep learning networks as ANNs owing to their analog nature of inputs and computations. This nomenclature helps to distinguish them from Spiking Neural Networks (SNNs) which perform event-based computations.}) based approaches for optical flow estimation have gained immense popularity \cite{dosovitskiy2015flownet,ren2017unsupervised}.
In general, these methods rely on the standard frame-based cameras as input sensors that capture pixel intensities over the entire frame at a regular sampling rate. However, the frame-based images suffer from a variety of issues such as motion blur and temporal aliasing when capturing high speed motion due to fixed low temporal resolution. They are also unable to perceive information accurately in high dynamic range scenes due to uneven exposure \cite{gallego2019event}. 

Event-based cameras, such as Dynamic Vision Sensors (DVS) \cite{dvs128}, address these problems by asynchronously sampling intensity changes on each pixel element, generating a stream of asynchronous events. This grants promising advantages, namely high temporal resolution (10$\mu$s vs 3ms), high dynamic range (140dB vs 60dB) and low power consumption (10mW vs 3W) compared to standard frame-based cameras \cite{gallego2019event}. Note, event cameras only capture the varying components of visual signals, generating sparse event streams. Hence, the output prediction becomes limited only at pixels-points where events exist, adding difficulty towards encoding the scene context. 

As is evident from the above discussions, none of the above sensors by themselves is able to effectively capture all relevant information of a scene. The limited applicability of each individual camera gives rise to the need for an optimal sensor-fusion technique, enabling the sensors to complement the limitations of each other. Such a technique would provide a practical solution towards accurately estimating dense pixel-wise motion in challenging scenarios such as rapid motion and high dynamic range environments.

Conventional computer vision and ANN-based methods are incompatible at handling the discrete and asynchronous event streams from event-based camera in their native form. This is due to the fact that these methods are generally designed for frame-based images, assuming regular frame rate and brightness consistency over the entire frames. In this regard, Spiking Neural Networks (SNNs), inspired from biological neuronal mechanisms, show a great promise for directly handling event-camera outputs. Moreover, SNNs perform event-based operations by carrying out the computations only at the arrival of the input events, exploiting the inherent sparsity of spatio-temporal event streams and thus enabling energy-efficient computations on specialized neuromorphic hardware such as Loihi from Intel Labs \cite{loihi2018} and TrueNorth from IBM \cite{merolla2014million}.


In this work, we propose a method for combining the advantages of regular frame-based images and a stream of asynchronous events. For this purpose, we present Fusion-FlowNet, a deep fused spiking-analog architecture for estimating optical flow that uses sensors of different modalities (standard frame-based images and asynchronous event streams). 
Our main contributions are as follows:
\begin{itemize}[leftmargin=*]
\item We propose Fusion-FlowNet architecture composed of a fusion of SNNs and ANNs for simultaneously processing event streams and frame-based images, leveraging their complementary sensing capabilities.
\item We present a Signed Integrate-and-Fire (SIF) neuron model for SNNs which can generate spike outputs with polarity (either positive or negative). In addition, we show that the SIF model coupled with a surrogate gradient method enables end-to-end learning in SNNs.
\item We show that Fusion-FlowNet outperforms the corresponding previous works in terms of optical flow estimation on the Multi-Vehicle Stereo Event Camera (MVSEC) dataset. Furthermore, we analyze that Fusion-FlowNet provides substantial savings in terms of network parameters and computational energy cost.
\end{itemize}


\section{Related Works} \label{sec:related_works}
Over the past few years, there have been major advancements towards optical flow estimation using event-cameras. Conventional computer vision algorithms have been adapted to encompass the asynchronous event stream from these sensors in \cite{aung2018,benosman2,gallego2018}. 
In ANN-based approaches, the event streams are essentially accumulated for fixed time intervals to generate synchronous frames. In EV-FlowNet~\cite{zhu2018ev}, the recent event counts as well as pixel-wise last timestamp information are encoded in a frame-based representation. However, this approach heavily suffers during rapid motion and in scenarios with dense localized events, resulting in loss of rich spatio-temporal information. Researchers in \cite{zhu2019unsupervised} proposed a 3D input representation of events interpolated in a 3D volume with time dimension comprising the input channels to retain the temporal fidelity. Nevertheless, this approach struggled to estimate the dense predictions in image regions with fewer events.


In general, SNNs provide advantages towards directly handling the asynchronous events and exploiting the inherent temporal information. Recently, authors in Spike-FlowNet \cite{10.1007/978-3-030-58526-6_22} aimed to overcome a noticeable drawback of SNNs -- namely the “spike vanishing" phenomenon where the number of spikes drastically reduce in the deeper layers, hindering learning. They proposed to effectively integrate SNNs and ANNs into a single network with the SNN layers enabling efficient event stream handling and the ANN layers addressing the spike vanishing problem. Note, since they used only the event streams as input, the predictions were limited to only the pixel locations containing non-zero number of events. Hence, estimating dense motion behavior was greatly limited. 

In contrast, researchers in \cite{pan2020single} presented a two-step approach to estimate optical flow by jointly using a set of events and a single frame-based image. They employed an optimization-based method to restore a sharp intensity image from the inputs, followed by ANN-based flow estimation methods \cite{liu2019selflow, sun2018pwc} on the restored frame-based image to generate an optical flow prediction. Contrary to \cite{pan2020single}, we explore an end-to-end learning approach that can directly process event streams and frame-based images for predicting final outputs while skipping the image restoration step. Moreover, our proposed method utilizes all available event streams as well as frame-based images within a time window, enabling accurate optical flow estimations over longer time windows.

\section{Method} \label{sec:method}
\subsection{Sensors and Input representation} \label{subsec:sensors}

\subsubsection{Frame-based Camera} \label{subsubsec:frame_sensors}
Frame-based cameras have been widely popular for computer vision applications. They provide dense and highly accurate pixel intensity information as frames over regular time intervals. In general, the frame intensity information is pivotal in various computer vision applications that require the high degree of accuracy such as face and object recognition \cite{masi2018deep}. Optical flow estimation using ANNs requires consecutive frame-based images to pass through separate input channels to the network. In our work, this input representation is utilized for the ANN part of Fusion-FlowNet (Sec.~\ref{subsec:arch}).

\subsubsection{Event-based Camera} \label{subsubsec:event_sensors}
Event-based cameras are novel vision sensors, emulating the functionality of biological retina cells \cite{mahowald1994silicon}. 
Event cameras transmit a stream of asynchronous events as the outcome of tracking intensity changes ($I$) at each pixel element, thereby capturing the relative motion of objects in the scene. Whenever the logarithmic intensity change at a pixel element surpasses a specified threshold ($\theta$), a discrete event is asynchronously generated as follows:
\begin{equation}
\|\log(I_{t+1}) - \log(I_{t})\| \geq \theta \tag{1}
\label{eq1}
\end{equation}
Event cameras provide the data in Address Event Representation (AER) format which incorporates a tuple $\{x, y, t, p\}$, comprising the pixel address ($x$ and $y$ locations), timestamp ($t$) and polarity of the intensity change ($p$). Here, each ON/OFF polarity corresponds to the increase or decrease in intensity of the pixel, respectively. 

Event cameras may not be generally suited for vision applications which need precise intensity information. However, their high temporal resolution and high dynamic range in addition to having low power consumption, make them ideal for usage on resource constrained platforms operating in challenging environments. Optical flow estimation is one such task which heavily suffers in such environments when realized using standard frame-based cameras and can greatly benefit with the usage of event-cameras. 

\begin{figure}[h]
\begin{center}
\includegraphics[width=0.49\textwidth]{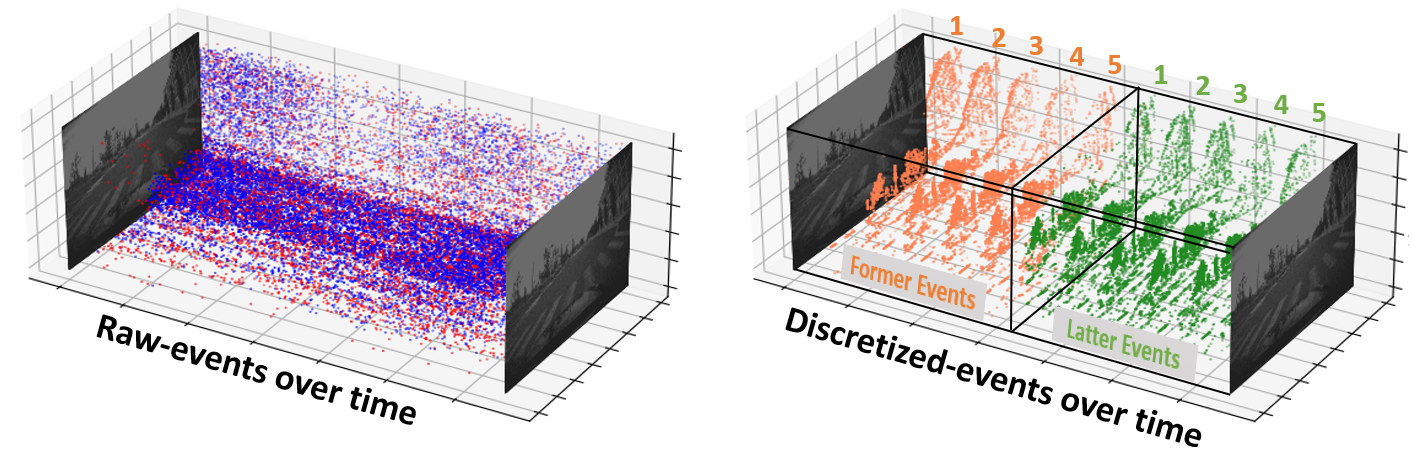}
\caption{($left$) Asynchronous raw event stream between two consecutive frame-based images. ($right$) Discretized event-frames between two consecutive frame-based images to shape the former and latter groups of events.}
\vspace{-4mm}
\label{fig:inp_rep}
\end{center}
\end{figure}

In our work, the raw event stream is transformed into two groups (former and latter) of discretized event frames and are passed as inputs to the SNN part of Fusion-FlowNet (Sec.~\ref{subsec:arch}). The input to the SNN encoder-branch consists of a sequence of event frames with four channels, each from the ON/OFF polarity of event frames from the former and the latter groups as illustrated in Fig.~\ref{fig:inp_rep}.
This representation preserves the spatio-temporal information in the event stream while displaying superior algorithmic performance and high energy-efficiency.

\subsubsection{Sensor-fusion} \label{subsubsec:sensor_fusion}
Interestingly, numerous available sensors, including the Dynamic and Active Vision Sensor (DAVIS)~\cite{dvs240}, are capable of simultaneously generating the asynchronous events as well as synchronous grayscale frames, simplifying the hardware costs of sensor-fusion. In addition, since there is a single camera coordinate system for both data modalities, the requirements for any expensive transformation and synchronization between multiple coordinate systems are eliminated. For this purpose, we employ the DAVIS sensor in this work.

In our work, the frame-based images serve two objectives. First, they are provided as network inputs and allow for dense optical flow predictions. Second, they are used for constructing the unsupervised loss required for training. On the other hand, the event streams are only provided as network inputs and enable accurate optical flow prediction in challenging environments as discussed previously. The proposed sensor fusion framework would thus allow to accurately estimate dense optical flow.

\subsection{Neuron Models} \label{subsec:neuron_models}
The primary difference between ANN and SNN operations is the notion of time. While ANNs feed-forward the dense analog-valued inputs at once, SNNs process the sparse binary inputs (spikes) as a function of time. Accordingly, different neuron models are employed in ANNs and SNNs.

\subsubsection{LeakyReLU Model} \label{subsubsec:leakyrelu}

In ANNs, LeakyReLU neuron \cite{xu2015empirical} replaces the negative part of the popular ReLU model by a linear function with a relatively small slope as below:
\[y =\begin{cases}
               x, & \text{if~$x>0$}\\
              \alpha x, & \text{otherwise}\\
            \end{cases} \tag{2}\]
where $\alpha$ is typically set to 0.01-0.1. Note, ReLU has a “dead neuron” problem that some neurons could get stuck in the negative side and play no role in discriminating between inputs. LeakyReLU addresses this problem by having a non-zero slope in the negative direction. This makes it useful especially for hard regression tasks such as motion estimation and predicting pixel-wise and high resolution outputs. In our work, LeakyReLU is employed for the ANN part of Fusion-FlowNet.

\begin{figure}[h]
\begin{center}
\includegraphics[width=0.47\textwidth]{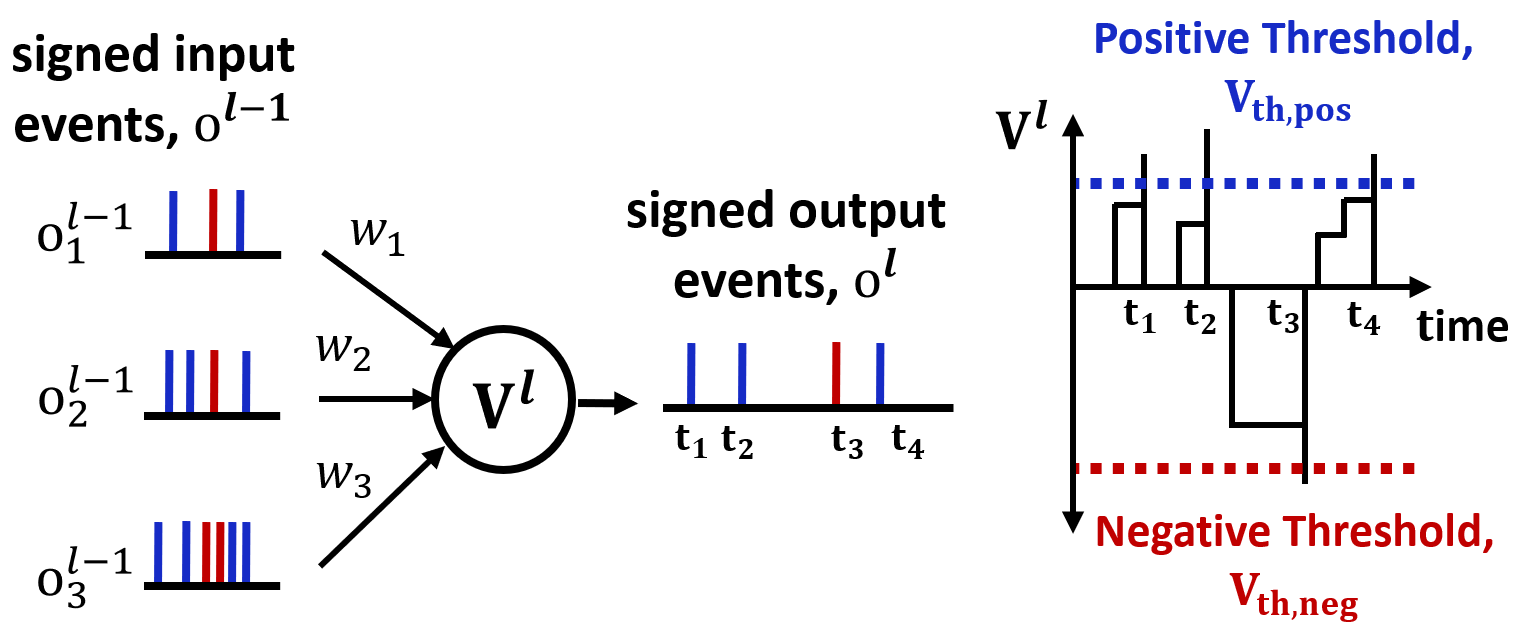}
\caption{Dynamics of Signed Integrate-and-Fire (SIF) neuron model. Whenever the membrane potential crosses either positive- or negative-threshold, the neuron fires a signed spike output and resets its membrane potential.}
\label{fig:sif_dynamics}
\vspace{-3mm}
\end{center}
\end{figure}

\begin{figure*}[h]
\begin{center}
\includegraphics[width=0.96\textwidth]{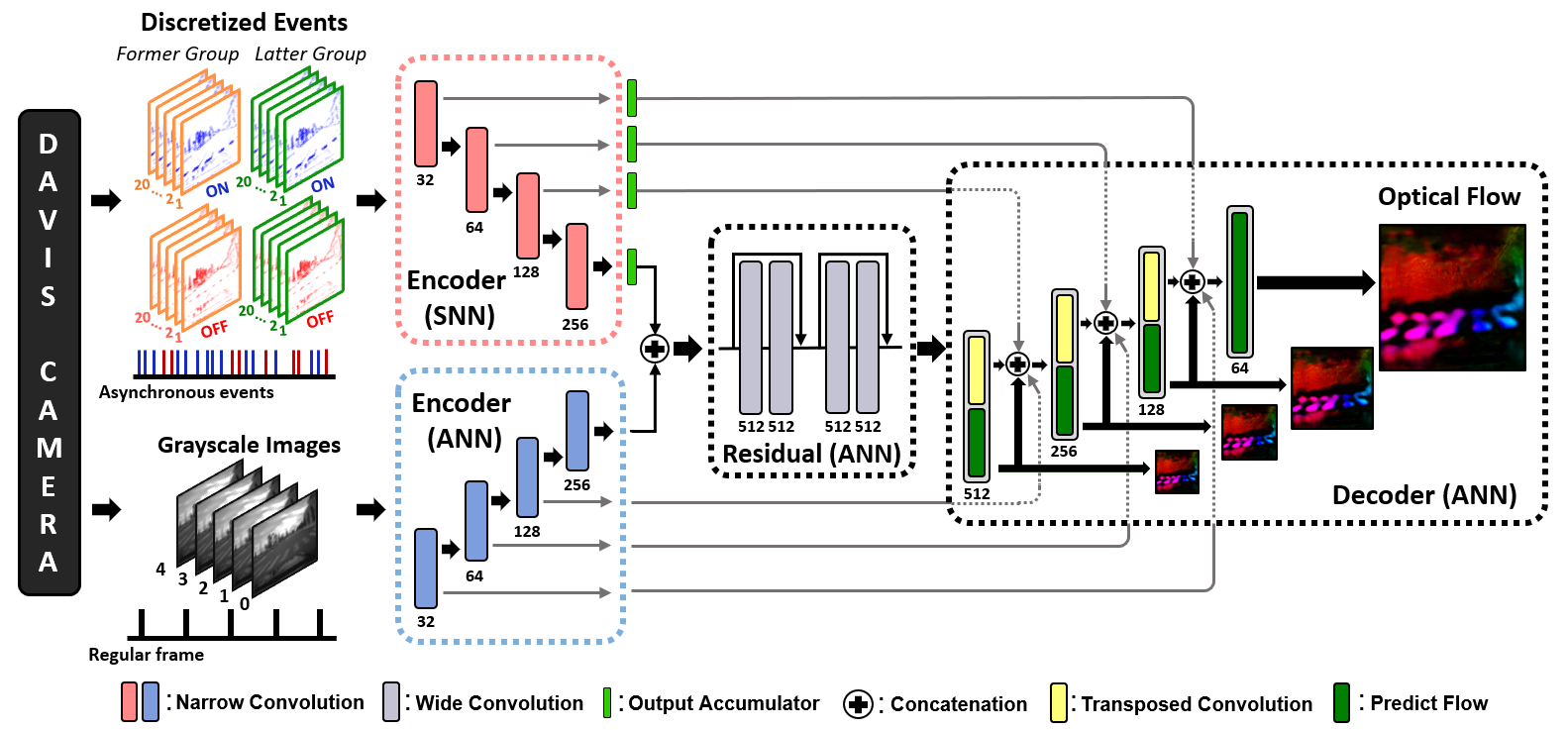}
\caption{Detailed illustration of Fusion-FlowNet$_{\text{Early}}$. The network contains the SNN- and ANN-based encoder-branches to extract features from asynchronous event streams and synchronized grayscale images, respectively. The rest of networks, involving residual and decoder blocks, are composed of ANN layers. The colors represent the types of layers. Best viewed in color.} 
\label{fig:arch}
\vspace{-1mm}
\end{center}
\end{figure*}

\subsubsection{Signed Integrate-and-Fire (SIF) Model} \label{subsubsec:sif}

Spiking neurons are inspired by biological models for emulating the efficient event-based operations in the human brain. In the literature, the Integrate-and-Fire (IF) neuron model \cite{burkitt2006review} is widely used for building SNNs because of its simplicity. In an IF neuron, input spikes are modulated by weight ($w$) and accumulated in an internal state of the neuron, called membrane potential over time. In the discrete time model, whenever the membrane potential ($v$) crosses a firing threshold, the neuron emits a binary output ($1$ or $0$) and resets the membrane potential as follows,  
\begin{equation}
v^l[n+1] = v^l[n] + w^{l}o^{l-1}[n] \tag{3}
\label{eq3}
\end{equation}
where $o^{l-1}[n]$ indicates the spike output from previous layer at time-step $n$. However, the IF neuron would also suffer from the “dead neuron” problem. To address this issue, we propose a Signed Integrate-and-Fire (SIF) neuron model that can generate signed spike outputs. The SIF neuron is equipped with positive and negative thresholds that enable the generation of positive- and negative-valued spike outputs, respectively. This operation is illustrated in Fig.~\ref{fig:sif_dynamics} and formulated as follows: 
\[o^{l} =\begin{cases}
               +1, & \text{if~$v^l>v_{th,pos}$}\\
               -1, & \text{elif~$v^l<v_{th,neg}$}\\
               0, & \text{otherwise}
            \end{cases} \tag{4}\]
However, the discontinuous and non-differentiable spike generation function of SIF model poses a critical challenge for conventional gradient-based learning. To overcome this challenge, we propose a surrogate gradient method for the SIF neuron to enable end-to-end backpropagation (discussed in Sec.~\ref{subsec:backprop}).

\subsection{Fusion-FlowNet Architecture} \label{subsec:arch}
The Fusion-FlowNet incorporates a deep fused network architecture that supports an end-to-end learning. It is built upon the U-Net architecture \cite{unet} that contains four encoder layers, two residual blocks and four decoder layers. The distinctions in our work involve the addition of dual pathways starting at the encoder, namely the SNN- and ANN-based branches. Each branch is composed of narrow convolution layers (similar to grouped convolutions used in AlexNet \cite{krizhevsky2012imagenet}) containing half the number of intermediate feature maps, compared to the original wide convolution layers. This is possible because of the usage of different modalities of input data, leading to reduction in network parameters without compromising on qualitative performance.

In the SNN-based encoder-branch, the four-channeled input event frames sequentially pass through the narrow convolution layers consisting of SIF neurons over time while being downsampled at each layer. At every time-step, the weighted spike outputs from each layer are integrated into the corresponding output accumulator. After passing all consecutive event images, the accumulated output is passed on ahead to subsequent layers.

In the ANN-based encoder-branch, the consecutive frame-based images in the time window pass through the narrow ANN layers in a single time step. Each ANN layer comprises of a convolution, batch-norm \cite{ioffe2015batch} and a LeakyReLU layer. Here too, the feature maps are downsampled at each layer.

After completing the forward propagation in both encoder-branches, the outputs are fused together before passing through the rest of the network. This is achieved by concatenating the intermediate activations from both the SNN and ANN branches at the same spatial locations. The fused activations from the last encoder layer pass through the residual blocks while the fused intermediate encoder outputs serve as input to corresponding layers in the decoder block. The four layers in the decoder block perform upsampling using transposed convolutions as well as produce multi-scale optical flow predictions. The multi-scale flow predictions, the transposed convolution outputs and the corresponding activations from the encoder layers are all concatenated together to construct the input for the next decoder layer. Finally, a full-scale optical flow prediction having the same dimension as the input frames is made at the final decoder layer. Fig.~\ref{fig:arch} showcases the proposed Fusion-FlowNet architecture, illustrating the discussed operations. 

\subsection{Unsupervised Training Method} \label{subsec:training}
Due to the limited availability of event-camera datasets containing ground-truth labels, we adopt an unsupervised approach to train the optical flow estimation network \cite{jason2016back}. Fusion-FlowNet is trained using unlabeled sequences, utilizing frame-based images for computing the loss. The overall loss function is composed of two parts: 
\begin{equation}
l_{\text{total}} = l_{\text{photo}} + \lambda l_{\text{smooth}} \tag{5}
\label{eq5}
\end{equation}
where $l_{photo}$ and $l_{smooth}$ represent photometric loss and smoothness loss respectively, and $\lambda$ denotes the loss weight factor.

\subsubsection{Photometric Loss} \label{subsubsec:photoloss}
Photometric loss helps to realize the object motion over time by tracking the pixel intensities between images. It is computed by using the start and end-frame grayscale images $(I_t(x,y)$, $I_{t+dt}(x,y))$  as well as the predicted optical flow. A spatial transformer \cite{jaderberg2015} inversely warps the end-frame image $(I_{t+dt}(x,y))$ using the current estimated optical flow ($u,v$) to obtain an image prediction $(I_{t+dt}(x+u,y+v))$. Then, the photometric loss ($l_{photo}$) aims to minimize the discrepancy between the start-frame image $(I_{t}(x,y))$ and the image prediction $(I_{t+dt}(x+u,y+v))$. The computation is as follows:
\begin{multline}
l_{photo} = \sum_{x,y} \rho(I_t(x,y) - I_{t+dt}(x+u, y+v))  \tag{6}
\label{eq6}
\end{multline}
where $I_t$ $(I_{t+dt})$ indicates the pixel intensity of the first (last) frame-based image, $u$, $v$ are the flow estimates in the $x, y$ directions, $\rho$ is the robust Charbonnier loss $\rho(x) = (x^2 + \eta^2)^r$ used for outlier rejection \cite{sun2014}. We set $r$ = $0.45$ and $\eta$ = $1e^{-3}$ as they show optimal results in prior works \cite{zhu2018ev,10.1007/978-3-030-58526-6_22}.

\subsubsection{Smoothness Loss} \label{subsubsec:smoothloss}
Smoothness loss ($l_{smooth}$) is applied to reduce the optical flow deviations between neighboring pixels by adding a regularizing effect on the predicted flow. It is computed as follows:
\begin{multline}
    l_{smooth}  = \sum_{j} \sum_{i} (\|u_{i,j}-u_{i+1,j}\|  + \|u_{i,j}-u_{i,j+1}\| \\+ \|v_{i,j}-v_{i+1,j}\| + \|v_{i,j}-v_{i,j+1}\|)  \tag{7}
\label{eq7}
\end{multline}
where $u_{i,j}$ and $v_{i,j}$ are the flow estimates at pixel location ($i,j$) in the $x$ and $y$ directions, respectively.

\subsection{Backpropagation in Fusion-FlowNet} \label{subsec:backprop}
\begin{figure}[h]
\begin{center}
\includegraphics[width=0.47\textwidth]{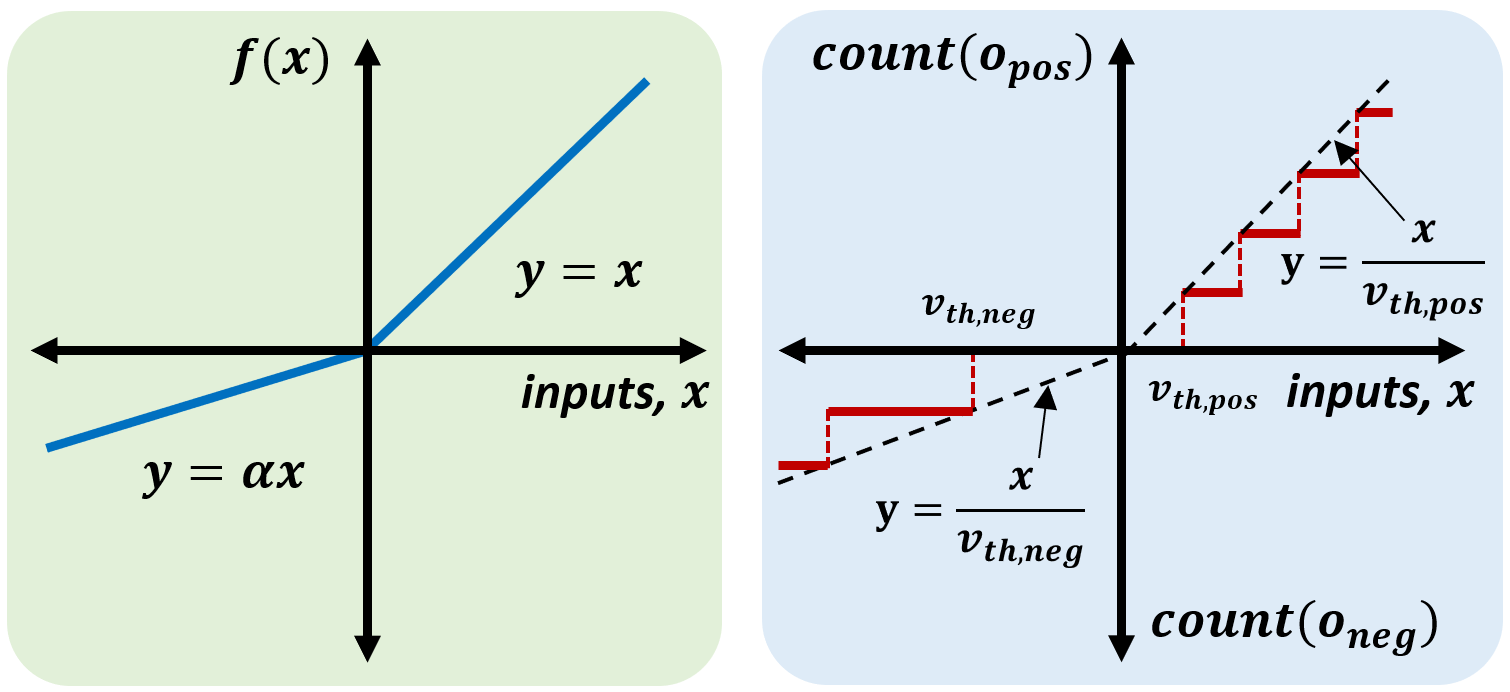}
\caption{Illustration of activation functions ($left$) LeakyReLU neuron ($right$) Signed Integrate-and-Fire (SIF) neuron.}
\label{fig:neuron_activation}
\end{center}
\vspace{-3mm}
\end{figure}

After forward propagation, the final loss ($l_{total}$) is evaluated and used to perform the backward propagation of the gradients. In ANN layers, the LeakyReLU is a differentiable activation that can be represented by the linear functions where the slope differs in positive and negative parts of input as shown in $left$ of Fig.~\ref{fig:neuron_activation}. The derivative of LeakyReLU activation ($\pderiv{f(x)}{x}$) is unity when input is positive, $\alpha$ when input is negative, and zero otherwise. Hence, standard backpropagation can calculate the gradient of the loss function with respect to each weight using chain rule. The parameter updates for the $l^{th}$ ANN layer are described as follows: \vspace{-1mm}
\begin{equation}
\triangle w^{l}_{ANN} = \pderiv{\text{loss}}{f(x^l)} \pderiv{f(x^l)}{o^l} \pderiv{o^l}{w^l} \tag{8}
\label{eq8}
\end{equation}

By contrast, the spike generation mechanism of SIF neuron results in a hard threshold function, making it discontinuous and non-differentiable. Hence, standard backpropagation cannot be directly applied to SNNs in its native form as illustrated in $right$ of Fig.~\ref{fig:neuron_activation}. To overcome this impediment, we present a surrogate gradient method for approximately estimating the spike generation function of SIF neuron. The surrogate gradient of SIF model is herein computed as follows: 
\[\pderiv{o[n]}{v[n]} =\begin{cases}
               \frac{1}{V_{th,pos}}, & \text{if~$v^l>v_{th,pos}$}.\\
               \frac{1}{V_{th,neg}}, & \text{if~$v^l<v_{th,neg}$}.\\
               0, & \text{otherwise.}
            \end{cases} \tag{9}\]
where each threshold ($V_{th,pos}, V_{th,neg}$) accounts for the change in the signed spike outputs with respect to the inputs.

During the backward pass, the errors ($\pderiv{l_{total}}{o^{l}}$) are backpropagated through the SNN layers using the surrogate gradient method and BackPropagation Through Time (BPTT) \cite{werbos1990backpropagation}. In BPTT, the network is unrolled for all time-steps and the weight update is assessed as the sum of gradients over each time-step. The parameter updates of the $l^{th}$ SNN layer are described as follows: \vspace{-1mm}
\begin{equation}
\triangle w^{l}_{SNN} = \sum_{n} \pderiv{\text{loss}}{o^l[n]} \pderiv{o^l[n]}{v^l[n]} \pderiv{v^l[n]}{w^l} \tag{10}
\label{eq10}
\end{equation}

\begin{figure*}[ht]
\begin{center}
\includegraphics[width=0.95\textwidth]{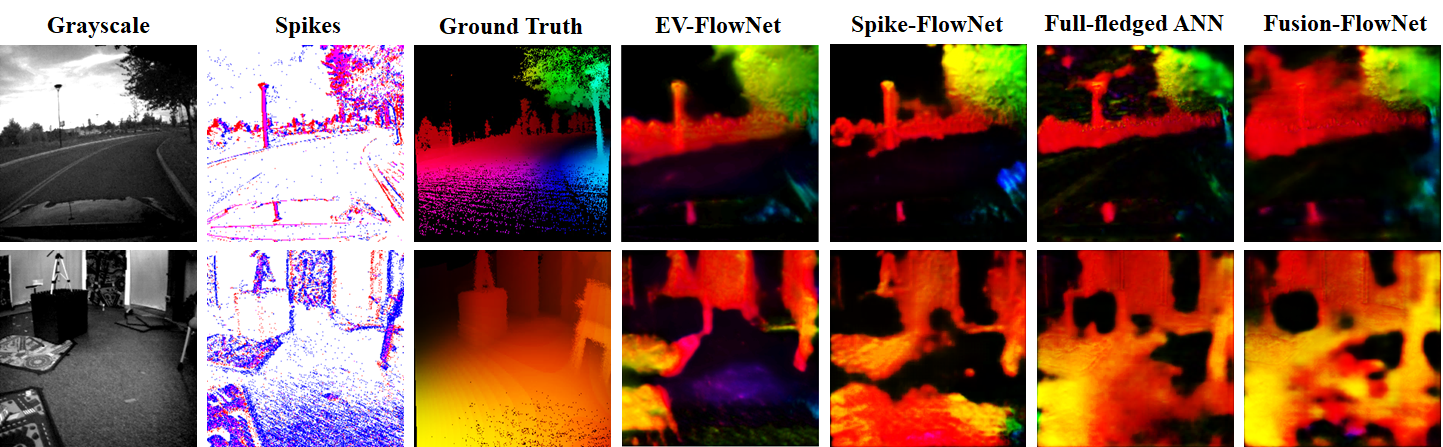}
\caption{Predicted optical flow compared with other state-of-the-art methods. EV-FlowNet~\cite{zhu2018ev} and Spike-FlowNet~\cite{10.1007/978-3-030-58526-6_22} use only the event stream as input. Full-fledged ANN uses only grayscale images as input. Fusion-FlowNet uses both the event stream as well as grayscale images. The samples are taken from ($top$) $outdoor\_day1$ and ($bottom$) $indoor\_flying2$. Best viewed in color.}
\label{fig:results}
\end{center}
 \vspace{-4mm}
\end{figure*}

\section{Experiments} \label{sec:experiments}

\subsection{Dataset and Training Details} \label{subsec:dataset_details}
We validate Fusion-FlowNet on the MVSEC dataset~\cite{zhu2018multivehicle} which contains events as well as grayscale frame sequences recorded using the DAVIS346 camera \cite{dvs240} in multiple indoor and outdoor environments. We use three indoor\_flying sequences and two outdoor\_driving sequences. The indoor\_flying sequences were collected using a drone flying in a closed room containing a variety of objects and are used mainly for evaluation. The outdoor\_day sequences were recorded from a car driving on public roads. We employ the outdoor\_day2 sequence for training and outdoor\_day1 sequence for evaluation. The training and evaluation are performed for two different time-window lengths (i.e, dt=1 and dt=4). Every consecutive pair of grayscale images encapsulate an event volume where dt=1 corresponds to constructing inputs using one such event volume while dt=4 corresponds to using four such volumes.

The event streams and frame-based images from left-camera are used for training. They are pre-processed by randomly cropping to $256$$\times$$256$ size and flipping horizontally and vertically (with $0.5$ probability). The learning rate is scaled by $0.7$ every $5$ epochs until $20$ epoch, and every $10$ epochs thereafter. The number of event frames in each group are set to $5$ for the $dt=1$ case and $20$ for the $dt=4$ case. In ANN layers, LeakyReLU model is employed with an $\alpha$ of $0.1$. In SNN layers, the positive and negative thresholds of the SIF neuron are set to $0.75$ and $7.5$, respectively. The loss weight factor $\lambda$ is set to $0.0003$. 

\begin{table}[h]
\caption{$\text{AEE}_{event}$ comparison with previous works}
 \vspace{-2mm}
\begin{center}
\resizebox{0.48\textwidth}{!}{
\begin{tabular}{llcccccccccc}
\hline
  & & \multicolumn{4}{c}{dt=1 frame} &  & \multicolumn{4}{c}{dt=4 frame} \\ \cline{3-7} \cline{8-11} \vspace{-3mm}\\
  $\text{AEE}_{event}$ & & ind1  & ind2 & ind3 & out1 &  & ind1  & ind2 & ind3 & out1 \\ \cline{1-1}  \cline{3-6} \cline{8-11} \vspace{-3mm} \\
Zhu et al.'19            &  &  0.58     & 1.02    & 0.87    & \textbf{0.32}     &  & 2.18     & 3.85    & 3.18    & 1.30     \\ \vspace{-3.5mm} \\
EV-FlowNet              & &  1.03     & 1.72    & 1.53    & 0.49     &  & 2.25     & 4.05    & 3.45    & 1.23     \\ \vspace{-3.5mm} \\
Spike-FlowNet              & &  0.84     & 1.28    & 1.11    & 0.49     &  & 2.24     & 3.83    & 3.18    & \textbf{1.09}    \\ \vspace{-3.5mm}  \\
Fusion-FlowNet             &  &  \textbf{0.56}     & \textbf{0.95}    & \textbf{0.76}    & 0.59     &  & \textbf{1.68}     & \textbf{3.24}    &\textbf{2.43}    & 1.17    \\ \vspace{-3.5mm}  \\\hline \vspace{-5mm}\\
\end{tabular}}
\end{center}
\vspace{-2mm}
\label{table1}
\end{table}

\subsection{Evaluation of Optical Flow} \label{subsec:eval}
For evaluation, the center cropped images of $256$$\times$$256$ size are taken from indoor\_flying1,2,3 and outdoor\_day1 sequences. For indoor\_flying sequences, events and grayscale frames corresponding to the entire sequences are used for evaluation. However, for outdoor\_day1 sequence, 800 grayscale frames and the asscociated event streams are used for evaluation as suggested in \cite{10.1007/978-3-030-58526-6_22,zhu2018ev}. For quantitative results, we calculate the standard Average End-point Error (AEE) which is the mean Euclidean distance between the estimated flow ($y_{\text{estim}}$) and the provided ground-truth ($y_{\text{gt}}$). In our work, we measure the two types of AEE results: (1) over all pixels ($\text{AEE}_{all}$) and (2) over pixels where events are present within the time-window ($\text{AEE}_{event}$).
\begin{equation}
\text{AEE} = \frac{1}{m} \sum_{\text{m}} \left\Vert(u,v)_{\text{estim}}-(u,v)_{\text{gt}}\right\Vert_2 \tag{11}
\label{eq11}
\end{equation}
\noindent where $m$ indicates the count of active pixels in the event frames for $\text{AEE}_{event}$ and every pixels of images for $\text{AEE}_{all}$.

\begin{table*}[h]
\caption{Average Endpoint Error (AEE) results for ablation studies}
 \vspace{-2mm}
\begin{center}
\resizebox{1\textwidth}{!}{
\begin{tabular}{ccccccccccccccccccccc}
\hline \vspace{-3.5mm}\\ \vspace{0.5mm}
                                 &  & \multicolumn{4}{c}{indoor1}                         &  & \multicolumn{4}{c}{indoor2}                         &  & \multicolumn{4}{c}{indoor3}                         &  & \multicolumn{4}{c}{outdoor1}                        \\ \cline{3-6} \cline{8-11} \cline{13-16} \cline{18-21} \vspace{-3.4mm}\\ 
                                 &  & \multicolumn{2}{c}{dt=1} & \multicolumn{2}{c}{dt=4} &  & \multicolumn{2}{c}{dt=1} & \multicolumn{2}{c}{dt=4} &  & \multicolumn{2}{c}{dt=1} & \multicolumn{2}{c}{dt=4} &  & \multicolumn{2}{c}{dt=1} & \multicolumn{2}{c}{dt=4} \\ \cline{3-6} \cline{8-11} \cline{13-16} \cline{18-21} \vspace{-3.6mm}\\ \vspace{0.1mm}
                                 &  & event        & all           & event        & all        &  & event        & all           & event        & all        &  & event        & all           & event        & all        &  & event        & all          & event        & all       \\ \cline{1-1} \cline{3-6} \cline{8-11} \cline{13-16} \cline{18-21} \vspace{-3mm} \\

Fusion-FlowNet$_{\text{Early}}$  &  &     \textbf{0.56}      & \textbf{0.62}         & \textbf{1.68}         &  \textbf{1.81}         &  &   \textbf{0.95}        & \textbf{0.89}         & \textbf{3.24}         &   \textbf{2.90}        &  &    \textbf{0.76}      & \textbf{0.85}         & \textbf{2.43}         &   \textbf{2.46}        &  &     0.59       & 1.02        & 1.17         &     \textbf{3.06}      \\
Fusion-FlowNet$_{\text{Late}}$  &  &     0.57      & 0.63         & 1.71         &  1.89         &  &   0.99        & 0.92         & 3.26         &   2.93       &  &    0.79       & 0.87         & 2.46         &   2.54        &  &     0.55       & 1.00        & 1.34         &   3.48      \\ \vspace{-3.5mm} \\ 
\cline{1-1} \cline{3-6} \cline{8-11} \cline{13-16} \cline{18-21} \vspace{-3mm}\\
Fusion$_{\text{Early}}$ [IF model]  &  &     \textbf{0.56}      & \textbf{0.62}          & 1.72         &  1.93        &  &   0.97        & 0.90         & 3.36         &   3.07        &  &    0.78       & 0.87         & 2.51         &   2.63        &  &     0.58       & 1.04        & 1.37         &     3.52      \\
Fusion$_{\text{Late}}$ [IF model] &  &     0.57     & 0.64         & 1.71         &  1.90         &  &   1.00        & 0.93         & 3.41         &   3.08        &  &    0.80       & 0.88         & 2.56         &   2.64        &  &     0.55       & \textbf{0.99}        & 1.38         &   3.53  \\ \vspace{-3.5mm}\\
\cline{1-1} \cline{3-6} \cline{8-11} \cline{13-16} \cline{18-21} \vspace{-3mm}\\ 
Spike-FlowNet  &  &     0.84      & 0.91         & 2.24         &  2.94         &  &   1.28        & 1.23         & 3.83         &   4.09        &  &    1.11       & 1.20         & 3.18         &   3.92        &  &     \textbf{0.49}       &  1.42        & \textbf{1.09}         &   3.28        \\
Full-fledged ANN  &  &     0.60      & 0.68        & 1.73        &  1.90        &  &   1.00        & 0.97         & 3.35         &   3.03        &  &    0.83       & 0.97         & 2.52         &   2.62        &  &     0.83       & 1.53        & 1.27         &  3.19 \\ \vspace{-4mm}\\ \cline{1-21} \vspace{-10mm} \\

\end{tabular}}
\label{table2}
\end{center}
\end{table*}

\subsection{Results} \label{subsec:results}
We compare Fusion-FlowNet with previous state-of-the-art works \cite{zhu2018ev,zhu2019unsupervised,10.1007/978-3-030-58526-6_22} in terms of the performance of optical flow prediction. As listed in Table~\ref{table1}, only $\text{AEE}_{event}$ results are compared here since other works do not provide AEE values for dense optical flow estimation. We observe that Fusion-FlowNet outperforms other implementations in almost all scenarios. The outdoor\_day1 sequence is known to have suffered from certain issues with its grayscale images during dataset creation, leading to anomalous results for $\text{AEE}_{event}$ as well as $\text{AEE}_{all}$. We report the results for it to maintain completeness in terms of comparison with previous works. Fig.~\ref{fig:results} visualizes the predicted flow for this work and compares it with previous state-of-the-art methods. 

\subsection{Ablation studies} \label{subsec:ablations}
\subsubsection{Architectural Variations} \label{subsubsec:arch_variations}
\begin{figure}[h]
\begin{center}
\includegraphics[width=0.5\textwidth]{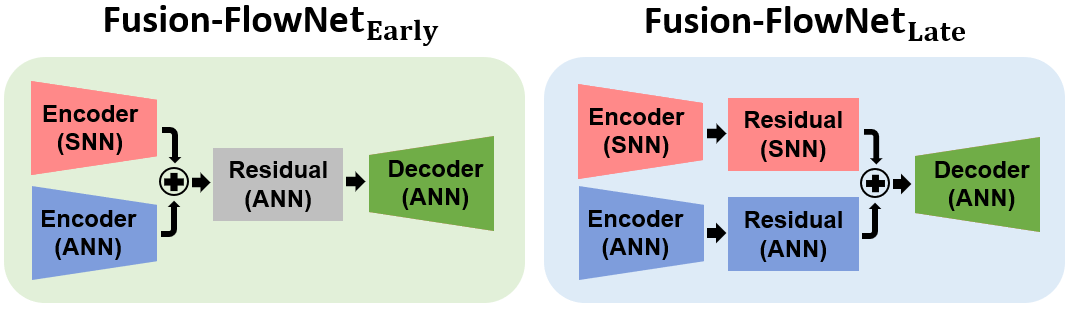}
\caption{Architectures of ($left$) Fusion-FlowNet$_{\text{Early}}$ and ($right$) Fusion-FlowNet$_{\text{Late}}$. Best viewed in color.}
\label{fig:early_late}
\vspace{-4mm}
\end{center}
\end{figure}

We perform an ablation study to analyze the effect of architectural variations on model performance and efficiency. We investigate a second architecture where the dual pathway branches are extended to residual blocks. As shown in Fig.~\ref{fig:early_late}, we denote the first architecture as Fusion-FlowNet$_{\text{Early}}$ and the second architecture as Fusion-FlowNet$_{\text{Late}}$. Rows $1$$-$$2$ in Table~\ref{table2} highlight the optical flow prediction capability of both the architectures. We find that Fusion-FlowNet$_{\text{Early}}$ outperforms Fusion-FlowNet$_{\text{Late}}$ in predicting accurate optical flow outputs.
Fusion-FlowNet$_{\text{Early}}$ contains comparatively larger number of parameters and fuses the intermediate features from the ANN/SNN branches in early layers, leading to better AEE results. On the other hand, Fusion-FlowNet$_{\text{Late}}$ performs the fusion at later layers leading to promising advantages in further reducing the network parameters and computational energy cost, as shown in Table~\ref{table3}.  

\subsubsection{Neuron Model Choice} \label{subsubsec:neuron_choice}
For investigating the benefits of proposed SIF neuron, we compare the variations of Fusion-FlowNet with the SNN blocks composed of SIF and IF neuron models. Rows $3$$-$$4$ in Table~\ref{table2} provide the AEE results for Fusion-FlowNet with IF neurons in the SNN layers. A comparison with results in rows $1$$-$$2$ show that networks using SIF model can predict more accurate flow outputs compared to networks using IF model. This establishes the benefit of the SIF model towards mitigating the “dead neuron” problem in deep SNN layers. 

\begin{table*}[h]
\begin{center}
\caption{Comparison of number of parameters and computational energy cost for different architectures for dt=1 and dt=4 cases. Lowest numbers highlighted in bold. (results averaged over all indoor and outdoor1 sequences)}
 \vspace{+2mm}
\resizebox{1\textwidth}{!}{
\begin{tabular}{ccccccccccccccccccc}
\hline  \vspace{-3mm}\\ \vspace{1mm}
                                 &  & \multicolumn{2}{c}{\#Parameters ($\times10^6$)} &  & \multicolumn{2}{c}{\text{\#OPS}$_{\text{ANN}}$ ($\times10^9$)} &  & \multicolumn{2}{c}{Spiking Activity ($\%$)} &  & \multicolumn{2}{c}{\text{\#OPS}$_{\text{SNN}}$ ($\times10^6$)} &  & \multicolumn{2}{c}{\text{E}$_{\text{Total}}$ (mJ)} &  &
                                 \multicolumn{2}{c}{Improvement}\\ \cline{3-4} \cline{6-7} \cline{9-10} \cline{12-13} \cline{15-16} \cline{18-19} \vspace{-3mm} \\ 
                                 &  & dt=1         & dt=4         &  & dt=1         & dt=4         &  & dt=1         & dt=4         &  & dt=1          & dt=4  &  & dt=1          & dt=4 &  & dt=1          & dt=4       \\ \cline{1-1} \cline{3-4} \cline{6-7} \cline{9-10} \cline{12-13}  \cline{15-16} \cline{18-19} \vspace{-3mm} \\
Full-fledged ANN                          &  & 13.044 &  13.046        &  & 5.339 & 5.367              &  &   -- & --        &  &   -- & --      &  &   24.536  & 24.666   &  &   1.00$\times$  & 1.00$\times$ \\ \vspace{-3.5mm} \\
Spike-FlowNet                      &  & 13.039 & 13.039         &  &  4.409 & 4.409             &  & 0.480 & 1.008      &  & 15.81 & 195.99     &  &   20.296 & 20.458    &  &   1.21$\times$  & 1.21$\times$ \\ \vspace{-3.5mm} \\
Fusion-FlowNet$\_{\text{Early}}$         &  & 12.269 & 12.270         &  & 4.648 & 4.648              &  & 0.173 & \textbf{0.174}      &  &  \textbf{1.03} & \textbf{4.18}   &  &   21.381  & 21.384   &  &   1.15$\times$  & 1.15$\times$ \\ \vspace{-3.5mm} \\
Fusion-FlowNet$\_{\text{Late}}$          &  & \textbf{7.549} & \textbf{7.550}           &  & \textbf{2.849} & \textbf{2.849}              &  & \textbf{0.147} & 0.179      &  & 5.24  & 6.44   &  &   \textbf{13.113}  & \textbf{13.114}   &  &   \textbf{1.87$\times$}  & \textbf{1.88$\times$} \\ \vspace{-3.5mm} \\ 
\hline \\ \vspace{-14mm}
\end{tabular}}
\label{table3}
\end{center}
\end{table*}

\subsubsection{Sensor Fusion} \label{subsubsec:sensor_fusion}
We study the usefulness of sensor fusion approach against single sensor approaches using inputs as either the event streams or frame-based images. For the event only approach, we investigate Spike-FlowNet \cite{10.1007/978-3-030-58526-6_22}, a hybrid neural architecture where the initial layers are composed of SNNs and the deeper layers are composed of ANNs. Note, Spike-FlowNet utilizes the similar event-based input representation scheme and unsupervised learning method, providing a fair comparison. For the frame-based image only approach, we implement a custom full-fledged ANN architecture that resembles the U-Net \cite{unet} architecture, and train it with the equivalent unsupervised method as Fusion-FlowNet.

Rows $5$$-$$6$ of Table~\ref{table2} summarize the results for the single sensor approaches. Unsurprisingly, both Fusion-FlowNet$_{\text{Early}}$ and Fusion-FlowNet$_{\text{Late}}$ achieve better AEE performances in dt=1 and dt=4 scenarios compared to single sensor approaches. This verifies that the proposed fusion approach benefits from utilizing the complementary characteristics of event- and frame-based images, leading to better performance in both slow- and fast-motion scenarios. Furthermore, in comparison to prior works as listed in Table~\ref{table1}, both fusion options provide superior AEE results. 

\subsection{Computational Efficiency} \label{subsec:energy}
We validate the efficiency of Fusion-FlowNet in terms of the number of network parameters and computational energy cost for inference. Table ~\ref{table3} provides a detailed analysis on computational efficiency along with the comparison with previously discussed alternate architectures.

We observe that both Fusion-FlowNet$_{\text{Early}}$ and Fusion-FlowNet$_{\text{Late}}$ contain fewer number of parameters compared to a full-fledged ANN architecture and Spike-FlowNet. This is due to the usage of narrow convolution layers which greatly reduce the number of parameters and computations. In particular, Fusion-FlowNet$_{\text{Late}}$ contains the least number of network parameters ($\sim58\%$ compared to full-fledged ANN) as the residual blocks contain the majority of the parameters and utilizing narrow convolutional layers for them helps reduce the total network parameters drastically.

For estimating the computational energy cost for different architectures, we first describe how computations in SNNs and ANNs differ from each other. Conceptually, SNNs perform highly sparse asynchronous ACcumulate (AC) operations over time. These synaptic operations are executed only at the arrival of input spikes due to the nature of binary-valued inputs. In contrast, ANNs perform expensive Multiply-and-ACcumulate (MAC) operations for computing dense Matrix-Vector Multiplications (MVMs). 
Based on the findings in \cite{horowitz20141}, a MAC operation requires a total of $E_{MAC}$=$4.6pJ$ of energy while an AC operation requires only $E_{AC}$=$0.9pJ$ for a 32-bit floating-point computation (45nm CMOS technology). This leads to the AC operation being $~5.1 \times$ more energy-efficient compared to the MAC operation. These findings coupled with the number of synaptic operations are commonly used to benchmark the computational energy cost of SNNs. \cite{merolla2014million, rueckauer2017conversion, 10.1007/978-3-030-58526-6_22}.

Next, we calculate the total number of synaptic operations for every layer. In the SNN layers, the number of synaptic operations are obtained by multiplying the pre-spike activities, the number of synaptic connections and the number of time-steps. Also, the computational energy of AC and MAC computations are taken into consideration for SNNs and ANNs, respectively. The total computational energy cost can be formalized as:
\begin{equation}
\text{\#OPS}_{\text{SNN}} =  N\sum_{l}M_l  C_l F_l, \;
\text{\#OPS}_{\text{ANN}} = \sum_{l} M_l C_l\\\tag{12}
\label{eq12}
\end{equation}
\vspace{-3mm}
\begin{equation}
\text{E}_{\text{Total}} = \text{\#OPS}_{\text{SNN}} \times \text{E}_{\text{AC}} + \text{\#OPS}_{\text{ANN}} \times \text{E}_{\text{MAC}}\\ \tag{13}
\label{eq13}
\end{equation}
where $M$ is the number of neurons, $C$ is the number of synaptic connections, $F$ represents mean spiking activity, $N$ is the number of timesteps, $\text{\#OPS}_{\text{SNN}}$/$\text{\#OPS}_{\text{ANN}}$ indicate the number of operations for SNN/ANN portions, and $\text{E}_{\text{Total}}$ denotes the total computational energy cost.

The last column in Table~\ref{table3} provides the overall improvement in computational energy cost. We observe that Fusion-FlowNet$_{\text{Late}}$ demonstrates the highest improvement in energy ($\sim1.88\times$) compared to full-fledged ANN. This is because more layers, including encoder and residual blocks, utilize narrow convolutions, leading to reduction in the number of parameters and consequently reduction in the total computational energy cost. Furthermore, the SNN pathway contributes negligibly to the total computational energy cost compared to ANN pathway.


\section{Conclusion} \label{sec:conclusion}
We propose a sensor/architecture fusion framework for accurately estimating optical flow in challenging environments. We leverage the complementary characteristics of event- and frame-based sensors as well as ANNs and SNNs. Our framework (Fusion-FlowNet) reports state-of-the-art optical flow prediction results, while substantially reducing network parameters and computational energy cost. This work contributes two different deep fused architectures (Fusion-FlowNet$_{\text{Early}}$ and Fusion-FlowNet$_{\text{Late}}$), having different applications of interest. Fusion-FlowNet$_{\text{Early}}$ provides highly accurate dense optical flow, proving to be appropriate for safety-critical applications. While, Fusion-FlowNet$_{\text{Late}}$ promises immense benefits in terms of computational efficiency, making it suitable for the edge applications on resource-constrained hardware.

{\small
\bibliographystyle{ieee_fullname}
\bibliography{egbib}

\begin{thebibliography}{10}\itemsep=-1pt

\bibitem{aung2018}
Myo~Tun Aung, Rodney Teo, and Garrick Orchard.
\newblock Event-based plane-fitting optical flow for dynamic vision sensors in
  fpga.
\newblock In {\em 2018 IEEE International Symposium on Circuits and Systems
  (ISCAS)}, pages 1--5, May 2018.

\bibitem{beauchemin1995computation}
Steven~S. Beauchemin and John~L. Barron.
\newblock The computation of optical flow.
\newblock {\em ACM computing surveys (CSUR)}, 27(3):433--466, 1995.

\bibitem{benosman2}
Ryad Benosman, Charles Clercq, Xavier Lagorce, Sio-Hoi Ieng, and Chiara
  Bartolozzi.
\newblock Event-based visual flow.
\newblock {\em IEEE Transactions on Neural Networks and Learning Systems},
  25(2):407--417, Feb 2014.

\bibitem{dvs240}
Christian Brandli, Raphael Berner, Minhao Yang, Shih-Chii Liu, and Tobi
  Delbruck.
\newblock A 240 × 180 130 db 3 µs latency global shutter spatiotemporal
  vision sensor.
\newblock {\em IEEE Journal of Solid-State Circuits}, 49(10):2333--2341, Oct
  2014.

\bibitem{burkitt2006review}
Anthony~N Burkitt.
\newblock A review of the integrate-and-fire neuron model: I. homogeneous
  synaptic input.
\newblock {\em Biological cybernetics}, 95(1):1--19, 2006.

\bibitem{loihi2018}
Mike Davies, Narayan Srinivasa, Tsung-Han Lin, Gautam Chinya, Yongqiang Cao,
  Sri~Harsha Choday, Georgios Dimou, Prasad Joshi, Nabil Imam, Shweta Jain,
  Yuyun Liao, Chit-Kwan Lin, Andrew Lines, Ruokun Liu, Deepak Mathaikutty,
  Steven McCoy, Arnab Paul, Jonathan Tse, Guruguhanathan Venkataramanan,
  Yi-Hsin Weng, Andreas Wild, Yoonseok Yang, and Hong Wang.
\newblock Loihi: A neuromorphic manycore processor with on-chip learning.
\newblock {\em IEEE Micro}, 38(1):82--99, January 2018.

\bibitem{dosovitskiy2015flownet}
Alexey Dosovitskiy, Philipp Fischer, Eddy Ilg, Philip Hausser, Caner Hazirbas,
  Vladimir Golkov, Patrick Van Der~Smagt, Daniel Cremers, and Thomas Brox.
\newblock Flownet: Learning optical flow with convolutional networks.
\newblock In {\em Proceedings of the IEEE international conference on computer
  vision}, pages 2758--2766, 2015.

\bibitem{gallego2019event}
Guillermo Gallego, Tobi Delbruck, Garrick Orchard, Chiara Bartolozzi, Brian
  Taba, Andrea Censi, Stefan Leutenegger, Andrew Davison, J{\"o}rg Conradt,
  Kostas Daniilidis, et~al.
\newblock Event-based vision: A survey.
\newblock {\em arXiv preprint arXiv:1904.08405}, 2019.

\bibitem{gallego2018}
Guillermo Gallego, Henri Rebecq, and Davide Scaramuzza.
\newblock A unifying contrast maximization framework for event cameras, with
  applications to motion, depth, and optical flow estimation.
\newblock {\em CoRR}, abs/1804.01306, 2018.

\bibitem{horowitz20141}
Mark Horowitz.
\newblock 1.1 computing's energy problem (and what we can do about it).
\newblock In {\em 2014 IEEE International Solid-State Circuits Conference
  Digest of Technical Papers (ISSCC)}, pages 10--14. IEEE, 2014.

\bibitem{ioffe2015batch}
Sergey Ioffe and Christian Szegedy.
\newblock Batch normalization: Accelerating deep network training by reducing
  internal covariate shift.
\newblock {\em arXiv preprint arXiv:1502.03167}, 2015.

\bibitem{jaderberg2015}
Max Jaderberg, Karen Simonyan, Andrew Zisserman, and koray kavukcuoglu.
\newblock Spatial transformer networks.
\newblock In C. Cortes, N. Lawrence, D. Lee, M. Sugiyama, and R. Garnett,
  editors, {\em Advances in Neural Information Processing Systems}, volume~28.
  Curran Associates, Inc., 2015.

\bibitem{jason2016back}
J~Yu Jason, Adam~W Harley, and Konstantinos~G Derpanis.
\newblock Back to basics: Unsupervised learning of optical flow via brightness
  constancy and motion smoothness.
\newblock In {\em European Conference on Computer Vision}, pages 3--10.
  Springer, 2016.

\bibitem{krizhevsky2012imagenet}
Alex Krizhevsky, Ilya Sutskever, and Geoffrey~E Hinton.
\newblock Imagenet classification with deep convolutional neural networks.
\newblock In {\em Advances in neural information processing systems}, pages
  1097--1105, 2012.

\bibitem{10.1007/978-3-030-58526-6_22}
Chankyu Lee, Adarsh Kosta, Alex~Zihao Zhu, Kenneth Chaney, Kostas Daniilidis,
  and Kaushik Roy.
\newblock Spike-flownet: Event-based optical flow estimation with
  energy-efficient hybrid neural networks.
\newblock In {\em European Conference on Computer Vision}, pages 366--382.
  Springer, 2020.

\bibitem{dvs128}
Patrick Lichtsteiner, Christoph Posch, and Tobi Delbruck.
\newblock A 128$\times$ 128 120 db 15 $\mu$s latency asynchronous temporal
  contrast vision sensor.
\newblock {\em IEEE Journal of Solid-State Circuits}, 43(2):566--576, Feb 2008.

\bibitem{liu2019selflow}
Pengpeng Liu, Michael Lyu, Irwin King, and Jia Xu.
\newblock Selflow: Self-supervised learning of optical flow.
\newblock In {\em Proceedings of the IEEE/CVF Conference on Computer Vision and
  Pattern Recognition}, pages 4571--4580, 2019.

\bibitem{lucaskanade}
Bruce~D. Lucas and Takeo Kanade.
\newblock An iterative image registration technique with an application to
  stereo vision.
\newblock In {\em Proceedings of the 7th International Joint Conference on
  Artificial Intelligence - Volume 2}, IJCAI'81, pages 674--679, San Francisco,
  CA, USA, 1981. Morgan Kaufmann Publishers Inc.

\bibitem{mahowald1994silicon}
Misha Mahowald.
\newblock The silicon retina.
\newblock In {\em An Analog VLSI System for Stereoscopic Vision}, pages 4--65.
  Springer, 1994.

\bibitem{masi2018deep}
Iacopo Masi, Yue Wu, Tal Hassner, and Prem Natarajan.
\newblock Deep face recognition: A survey.
\newblock In {\em 2018 31st SIBGRAPI conference on graphics, patterns and
  images (SIBGRAPI)}, pages 471--478. IEEE, 2018.

\bibitem{merolla2014million}
Paul~A Merolla, John~V Arthur, Rodrigo Alvarez-Icaza, Andrew~S Cassidy, Jun
  Sawada, Filipp Akopyan, Bryan~L Jackson, Nabil Imam, Chen Guo, Yutaka
  Nakamura, et~al.
\newblock A million spiking-neuron integrated circuit with a scalable
  communication network and interface.
\newblock {\em Science}, 345(6197):668--673, 2014.

\bibitem{narayana2013coherent}
Manjunath Narayana, Allen Hanson, and Erik Learned-Miller.
\newblock Coherent motion segmentation in moving camera videos using optical
  flow orientations.
\newblock In {\em Proceedings of the IEEE International Conference on Computer
  Vision}, pages 1577--1584, 2013.

\bibitem{pan2020single}
Liyuan Pan, Miaomiao Liu, and Richard Hartley.
\newblock Single image optical flow estimation with an event camera.
\newblock {\em arXiv preprint arXiv:2004.00347}, 2020.

\bibitem{ren2017unsupervised}
Zhe Ren, Junchi Yan, Bingbing Ni, Bin Liu, Xiaokang Yang, and Hongyuan Zha.
\newblock Unsupervised deep learning for optical flow estimation.
\newblock In {\em Thirty-First AAAI Conference on Artificial Intelligence},
  2017.

\bibitem{unet}
Olaf Ronneberger, Philipp Fischer, and Thomas Brox.
\newblock U-net: Convolutional networks for biomedical image segmentation.
\newblock {\em CoRR}, abs/1505.04597, 2015.

\bibitem{rueckauer2017conversion}
Bodo Rueckauer, Iulia-Alexandra Lungu, Yuhuang Hu, Michael Pfeiffer, and
  Shih-Chii Liu.
\newblock Conversion of continuous-valued deep networks to efficient
  event-driven networks for image classification.
\newblock {\em Frontiers in neuroscience}, 11:682, 2017.

\bibitem{sun2014}
Deqing Sun, Stefan Roth, and Michael~J. Black.
\newblock A quantitative analysis of current practices in optical flow
  estimation and the principles behind them.
\newblock {\em Int. J. Comput. Vision}, 106(2):115--137, Jan. 2014.

\bibitem{sun2018pwc}
Deqing Sun, Xiaodong Yang, Ming-Yu Liu, and Jan Kautz.
\newblock Pwc-net: Cnns for optical flow using pyramid, warping, and cost
  volume.
\newblock In {\em Proceedings of the IEEE conference on computer vision and
  pattern recognition}, pages 8934--8943, 2018.

\bibitem{wang2011action}
Heng Wang, Alexander Kl{\"a}ser, Cordelia Schmid, and Cheng-Lin Liu.
\newblock Action recognition by dense trajectories.
\newblock In {\em CVPR 2011}, pages 3169--3176. IEEE, 2011.

\bibitem{werbos1990backpropagation}
Paul~J Werbos.
\newblock Backpropagation through time: what it does and how to do it.
\newblock {\em Proceedings of the IEEE}, 78(10):1550--1560, 1990.

\bibitem{xu2015empirical}
Bing Xu, Naiyan Wang, Tianqi Chen, and Mu Li.
\newblock Empirical evaluation of rectified activations in convolutional
  network.
\newblock {\em arXiv preprint arXiv:1505.00853}, 2015.

\bibitem{zhu2018ev}
Alex Zhu, Liangzhe Yuan, Kenneth Chaney, and Kostas Daniilidis.
\newblock Ev-flownet: Self-supervised optical flow estimation for event-based
  cameras.
\newblock In {\em Proceedings of Robotics: Science and Systems}, Pittsburgh,
  Pennsylvania, June 2018.

\bibitem{zhu2018multivehicle}
Alex~Zihao Zhu, Dinesh Thakur, Tolga {\"O}zaslan, Bernd Pfrommer, Vijay Kumar,
  and Kostas Daniilidis.
\newblock The multivehicle stereo event camera dataset: An event camera dataset
  for 3d perception.
\newblock {\em IEEE Robotics and Automation Letters}, 3(3):2032--2039, 2018.

\bibitem{zhu2019unsupervised}
Alex~Zihao Zhu, Liangzhe Yuan, Kenneth Chaney, and Kostas Daniilidis.
\newblock Unsupervised event-based learning of optical flow, depth, and
  egomotion.
\newblock In {\em Proceedings of the IEEE Conference on Computer Vision and
  Pattern Recognition}, pages 989--997, 2019.

\end{thebibliography}
}

\end{document}